\documentclass[letterpaper, 10 pt, conference]{ieeeconf}  % Comment this line out if you need a4paper
\IEEEoverridecommandlockouts                              % This command is only needed if 
\usepackage{graphicx} % for pdf, bitmapped graphics files
\usepackage{amsmath} % assumes amsmath package installed
\usepackage{amssymb}  % assumes amsmath package installed
\usepackage{xcolor}
\usepackage{url}

\newcommand{\CStart}[0]{\begin{tabular}[c]{@{}c@{}}}
\newcommand{\CEnd}[0]{\end{tabular}}

\title{\LARGE \bf
ALET$^*$ (Automated Labeling of Equipment and Tools): A Dataset for Tool Detection and Human Worker Safety Detection
}

\author{Fatih Can Kurnaz$^{1}$, Burak Hocao\~glu$^{1}$, Mert Kaan Y\i{}lmaz$^{1}$, \.{I}dil S\"ulo$^{1}$, and Sinan Kalkan$^{1}$% <-this % stops a space
    \\ 
\thanks{* The world `alet' means `tool' in Turkish. }% <-this % stops a space
\thanks{$^{1}$ KOVAN Research Lab, Dept. of Computer Engineering, Middle East Technical University, Ankara, Turkey
        {\tt\small \{fatih.kurnaz, burak.hocaoglu, kaan.yilmaz, idil.sulo, skalkan\}@metu.edu.tr}}%
}

\begin{document}

\maketitle
\thispagestyle{empty}
\pagestyle{empty}

\begin{abstract}
Robots collaborating with humans in realistic environments need to be able to detect the tools that can be used and manipulated. However, there is no available dataset or study that addresses this challenge in real settings. In this paper, we fill this gap with a dataset for detecting farming, gardening, office, stonemasonry, vehicle, woodworking, and workshop tools. The scenes in our dataset are snapshots of sophisticated environments with or without humans using the tools. The scenes we consider introduce several challenges for object detection, including the small scale of the tools, their articulated nature, occlusion, inter-class invariance, etc. Moreover, we train and compare several state of the art deep object detectors (including Faster R-CNN, Cascade R-CNN, RepPoint, and RetinaNet) on our dataset. We observe that the detectors have difficulty in detecting especially small-scale tools or tools that are visually similar to parts of other tools. In addition, we provide a novel, practical safety use case with a deep network which checks whether the human worker is wearing the safety helmet, mask, glass, and glove tools. With the dataset, the code and the trained models, our work provides a basis for further research into tools and their use in robotics applications. The dataset, the trained networks, and all associated codes will be made available at: \url{https://github.com/metu-kovan/METU-ALET}.
\end{abstract}

%%%%%%%%%%%%%%%%%%%%%%%%%%%%%%%%%%%%%%%%%%%%%%%%%%%%%%%%%%%%%%%%%%%%%%%%%%%%%%%%
\section{INTRODUCTION}

The near future will see a cohabitation of robots and humans, where they will work together for performing tasks that are especially challenging, tiring or unergonomic for humans. This requires robots to have abilities for perceiving the humans, the task at hand, and the environment. An essential perceptual component for these abilities is the detection of objects, especially the tools.

\begin{figure}[hbt!]
    \centering
    \includegraphics[width=0.4\textwidth]{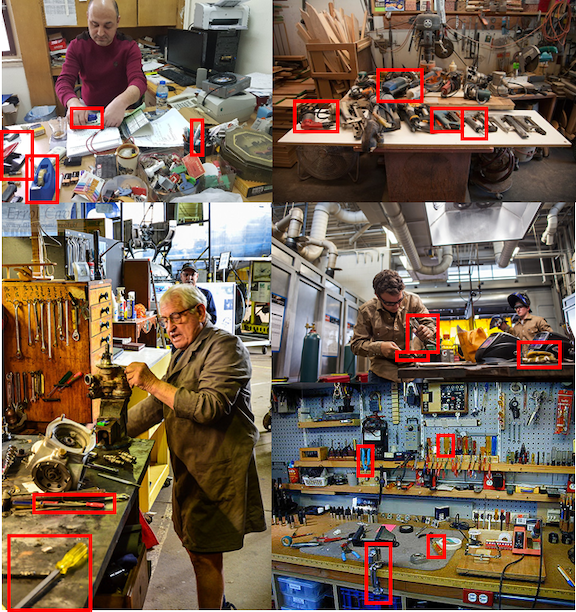}
    \caption{Samples from the METU-ALET dataset, illustrating the wide range of challenging scenes and tools that a robot is expected to recognize in a clutter, possible with human co-workers using the tools. Since annotations are too dense, only a small subset is displayed. [Best viewed in color] \label{fig:dataset_samples}}
\end{figure}

%Deep learning has made it possible to detect objects in challenging scenes with impressive performances. With the pioneering models such as Faster R-CNN \cite{faster_RCNN}, RetinaNet \cite{RetinaNet}, object detection has become an easy-to-integrate functionality for robotics applications.

The robotics community has paid marginal importance to tools that are used by humans. For example, there are studies focusing on affordances of tools, or on the detection and transfer of these affordances \cite{myers2015affordance,mar2018can,mar2018framework}. However, these studies considered tools mostly in isolated and limited, toy environments. Moreover, they have considered only a limited set of tools (see Table \ref{tab:datasets}). What is more, the literature has not studied detection of tools, nor is there a dataset available for it.

In this paper, we focus on the detection of tools in realistic, cluttered environments (e.g. like those in Fig.  \ref{fig:dataset_samples}) where collaboration between humans and robots is expected. To be more specific, we study detection of tools in real work environments that are composed of many objects (tools) that look alike and that occlude each other. For this end, we first collect an extensive tool detection dataset composed of 49 tool categories. Then, we compare the widely used state-of-the-art object detectors on our dataset, as a baseline. The results suggest that detecting tools is very challenging owing to tools being too small and articulated, and bearing too much inter-class similarity. Finally, we introduce a safety usecase from our dataset and train a novel CNN network for this task.

\textbf{The necessity for a dataset for tool detection:} Tool detection requires a dataset of its own since it bares novel challenges of its own: (i) Many tools are small objects which elicit a problem to standard object detectors that are tuned for detecting moderately larger objects. (ii) Many tools are articulated, and in addition to viewpoint, scale and illumination changes, object detectors need to address invariance to articulation. (iii) Tools are generally used in highly cluttered environments posing challenges on clutter, occlusion, appearance, and illumination -- see Figure \ref{fig:dataset_samples} for some samples. (iv) Many tools exhibit low inter-class differences (e.g., between screwdriver, chisel and file or between putty knives and scraper).

\begin{table*}[hbt]
    \centering
    \caption{A comparison of the datasets that include tools. The Epic-Kitchens dataset provides videos, which makes analysis of scenes difficult. Moreover, the figures for the Epic-Kitchens dataset are estimated based on the provided data. *SO denotes the number of scenes that only include a single object. \label{tab:datasets}}
    \scriptsize
    \begin{tabular}{ | c | c | c | c | c | c | c | } \hline
        \textbf{Dataset} & \CStart{} \textbf{Tool}\\ \textbf{Categories} \CEnd{}    &  \CStart{} \textbf{Tool}\\\textbf{Classes} \CEnd{}   & \CStart{} \textbf{\# of} \\\textbf{Images} \CEnd{} & \CStart{} \textbf{\# Instances}\\\textbf{per Tool} \CEnd{}  & \textbf{Modality} & \CStart{} \textbf{Dense}\\\textbf{Bounding Boxes?} \CEnd{} \\ \hline \hline
        \CStart{} RGB-D Part  Aff.  \cite{myers2015affordance}\CEnd{} & \CStart{} Kitchen, Workshop,\\ Garden \CEnd{} & 17 & \CStart{} 3\\(SO*: 102) \CEnd{} & 6.17 & RGB-D  & No \\ \hline
        %  &  Garden            &    & (SO*: 102)  &      &        & \\\hline
       \CStart{} ToolWeb  \cite{abelha2017learning}  \CEnd{} & \CStart{} Kitchen, Office, \\ Workshop \CEnd{} & 23 &  \CStart{} 0\\(SO*: 116)\CEnd{} & 5.03 & 3D model & No \\\hline
        %              &  Workshop          &    & (SO*: 116) &      &          & \\ \hline
       \CStart{} Visual Aff. of Tools \cite{dehban2016moderately} \CEnd{} & Toy & 3 & \CStart{} 0\\(SO*: 5280)\CEnd{} & 377  & \CStart{} RGB stereo \\ (inc. semantic map)\CEnd{}  & No \\ \hline
       %ImageNet \cite{IMAGENET} & \CStart{} Farm, Garden, Office,\\ Stonemasonry, Vehicles, \\ Woodwork, Workshop\CEnd{} & \CStart{}25\CEnd{} & ? & 200+ & RGB & No \\ \hline
       %of tools \cite{dehban2016moderately}           &                    &    & (SO*: 5280) & & (inc. semantic map) & \\\hline
       \CStart{} Epic-Kitchens \cite{Damen2018EPICKITCHENS} \CEnd{} & Kitchen & ~60+  & \CStart{}NA\\ (Videos)\CEnd{} & ~200+ & \CStart{} RGB stereo \\ (inc. semantic map)\CEnd{}             & Yes \\  \hline\hline
       %\cite{Damen2018EPICKITCHENS}         &         &      &   &      & (inc. semantic map) & \\\hline\hline
        \textbf{METU-ALET (Ours)} & \CStart{} Farm, Garden, Office,\\ Stonemasonry, Vehicles, \\ Woodwork, Workshop\CEnd{}  & 49 & \CStart{}2699\\(SO*: 0)\CEnd{} & 200+ & RGB  & Yes\\ \hline 
           %& Stonemasonry, Vehicles, &   & (SO*: 0) & & w bounding boxes & \\                & Woodwork, Workshop    & & & &  & \\\hline
    \end{tabular}
\end{table*}

\subsection{Related Work}
\label{sect:related_work}

%In this section, we review the related work, and provide a list of our contributions and a comparison with the literature.

\textbf{Object Detection:} Object detection is one of the most studied problems in Computer Vision with many practical applications in many robotics scenarios. Object detection generally follows a two-stage approach: (i) Region selection, which pertains to the selection of image regions that are likely to contain an object. (ii) Object classification, which deals with the classification of a selected region into one of the object categories. With advances in deep learning both stages have seen tremendous boost in performance in many challenging settings, e.g., \cite{RCNN, fast_RCNN, faster_RCNN}. Faster R-CNN \cite{faster_RCNN} is a well-known representative of such two-stage detectors.

It has been shown that the two stages can be combined and objects can be detected in one stage. Models such as \cite{RetinaNet,SSD,YOLO} assume a fixed set of localized image regions (anchors) for each object category, and estimate objectness for each category and for each anchor. Among these, RetinaNet \cite{RetinaNet} processes features and make classification at multiple scales (called a feature pyramid network) and combines the results, yielding the state of the art results among one-stage detectors.

These one-stage and two-stage detectors have a top-down approach. In contrast, bottom-up object detection forms a detection pipeline from points or keypoints that are likely to be on objects and identifies objects by combining such points (examples include CenterNet \cite{CenterNet}, RepPoints \cite{DBLP:RepPoint}).

The current trend in deep object detectors is to take a pre-trained object classification network as the feature extractor (called the backbone network), then perform two-stage or one-stage object detection from those features. Alternatives for the backbone network include deep classification networks such as VGG \cite{VGG}, ResNet \cite{RESNET} and ResNext \cite{RESNEXT}. 

\textbf{Tools in Robotics:} The robotics community has extensively studied how tools can be grasped, manipulated, and how such affordances can be transferred across tools. For example, Kemp and Edsinger \cite{kemp2006robot} focused on detection and the 3D localization of tool tips from optical flow. For the same goal, Mart et al. \cite{mar2018framework} proposed using a CNN network (AlexNet) to first classify a blob into one of the three tool labels they considered, and then used 3D geometric features to identify tool tips. 

Another study \cite{myers2015affordance} addressed the problem of estimating the grasping positions, scoops and supports of tools from RGB-D data. In a similar setting, Mar et al. \cite{mar2015multi,mar2018can} studied prediction of affordances for different categories of tools separately. For this purpose, they first clustered tools using their 3D geometric descriptors, and then estimated the affordances of each cluster separately.

%Another crucial problem in tools is the transfer of learned affordances across tools. For this, Abelha and Guerin \cite{abelha2017transfer} proposed a geometry-based reasoning where tools are segmented to semantically meaningful parts using 3D geometric features, and the affordances of the parts are shared across similar parts across different tools.

\textbf{Related datasets:} Comparing our dataset to the ones used in the robotics literature (e.g., \cite{myers2015affordance,abelha2017learning,dehban2016moderately} -- see also Table \ref{tab:datasets}), we see that they are limited in the number of categories and the instances that they consider. Moreover, since they mainly focus on detection of tool affordances, they are not suitable for training a deep object detector.

For related problems, there are numerous datasets of objects in the robotics literature, e.g. for 3D pose estimation and robot manipulation (e.g., LINEMOD \cite{hinterstoisser2012model}, YCB Objects \cite{calli2015benchmarking}, Table-top Objects \cite{sun2010depth}, Object Recognition Challenge \cite{vaskevicius2012jacobs}). These datasets generally include table-top objects with 3D models and do not include tool categories.

For object detection, there are several datasets such as PASCAL \cite{PASCAL}, MS-COCO \cite{MSCOCO} and ImageNet \cite{IMAGENET}, which do include some tool categories (e.g., hammer, scissors); however, these datasets are designed to be for general purpose objects and scenes, unlike the ones we expect to see in tool-used environments (such as the ones in Fig. \ref{fig:dataset_samples}). Therefore, they do not provide sufficient amount of training instances for training a general purpose tool detector with a reasonable performance.

\textbf{Safety Detection:} There are studies on detecting whether or not a human worker is wearing safety helmet \cite{helmet1,helmet2} or a vest \cite{helmet3}. However, there are no studies or datasets that include safety glass, mask, glove and headphone together with helmet.

\subsection{Contributions of Our Work}

The main contributions of our work can be summarized as follows:
\begin{itemize}
    \item \textbf{A Tool Detection Dataset}: To the best of our knowledge, ours is the first to provide a dataset on detection of tools in the wild. %As reviewed above, the robotics literature has focused on affordances of tools, 3D detection and pose estimation of objects, and neither set of studies provided a dataset on tools. The computer vision community, on the other hand, has extensive datasets for objects (e.g., PASCAL VOC, MS-COCO), which do not focus on tools. The Epic-Kitchen dataset focuses on object detection and action recognition in kitchen contexts, including also the tools that may be used in a kitchen.
    \item \textbf{A Baseline for Tool Detection}: On our dataset, we train and analyze many state-of-the-art deep object detectors. Together with the dataset, the code and the trained models, our work can form as a basis for robotics applications that require detection of tools in challenging realistic work environments with humans.
    \item \textbf{A Novel Usecase for Checking the Safety of Human Coworkers:} Our dataset includes humans performing tasks with and without wearing a safety helmet, mask, glass, headphone and gloves. We form a ALET Safety subset of positive and negative instances of these and train a deep CNN network that checks whether a human coworker is wearing these safety tools. %This usecase reinforces the practicality of our dataset.
\end{itemize}
\section{METU-ALET: A TOOL DETECTION DATASET}

In this section, we present and describe the details of METU-ALET and how the dataset was collected.

\subsection{Tools and Their Categories}
\label{sect:tools_and_categories}

In ALET, we consider 49 different tools that are used for six broad contexts or purposes: Farming, gardening, office, stonemasonry, vehicle, woodworking, workshop tools\footnote{It is better to call some of these objects as equipment. However, since they provide similar functionalities (being used by a human or a robot while performing a task), we will just use the term tool to refer to all such objects, for the sake of simplicity.}. The 20 most frequent tools from our dataset are:
Chisel, Clamp, Drill, File, Gloves, Hammer, Mallet, Meter, Pen, Pencil, Plane, Pliers, Safety glass, Safety helmet, Saw, Screwdriver, Spade, Tape, Trowel, Wrench.

We excluded tools used in kitchen since there is already an exclusive dataset for this purpose \cite{Damen2018EPICKITCHENS}. Moreover, we limited ourselves to tools that can be ultimately grasped, pushed, or manipulated in an easy manner by a robot. Therefore, we did not consider tools such as ladders, forklifts, and power tools that are bigger than a hand-sized drill.

\subsection{Dataset Collection}

\begin{figure}[hbt!]
    \centering
    \includegraphics[width=0.4\textwidth]{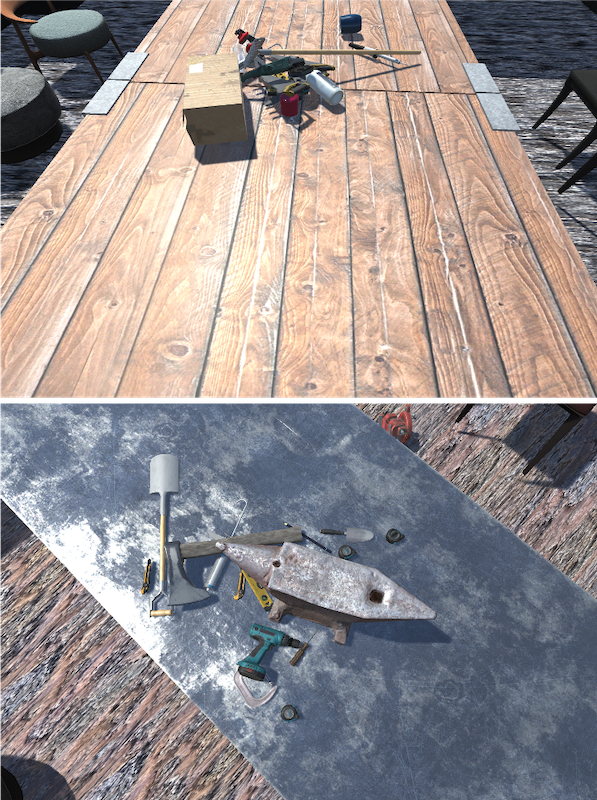}
    \caption{Some examples from the Synthetic Images. [Best viewed in color] \label{fig:synthetic_samples}}
\end{figure}

Our dataset is composed of three groups of images:

\begin{itemize}
    \item \textbf{Images collected from the web}: Using keywords and usage descriptions that describe the tools listed in Section \ref{sect:tools_and_categories}, we crawled and collected  royalty-free images from the following websites:  Creativecommons, Wikicommons, Flickr, Pexels, Unsplash, Shopify, Pixabay, Everystock, Imfree.
    
    \item \textbf{Images photographed by ourselves}: We captured photos of office and workshop environments from our campus.
    
    \item \textbf{Synthetic images}: 
    In order to make sure that there are at east 200 instances for each tool, we developed a simulation environment and collected synthetic images (see Fig. \ref{fig:synthetic_samples} for examples). For this, we used the Unity3D platform with 3D models of tools acquired from UnityStore. For each scene to be generated, the following steps are followed:
    \begin{itemize}
        \item \textit{Scene Background}: We created a room like environment with 4 walls, 10 different random objects (chair, sofa, corner-piece, television) in static positions. At the center of the room, we spawned one of six different tables selected randomly from $Uniform(1, 6)$. To introduce more randomness, we also dropped unrelated objects like mugs, bottles etc. randomly.
        \item \textit{Camera}: Each dimension of the camera position (x, y, z) was sampled randomly from $Uniform(-3, 3)$. Camera's viewing direction was set towards the center of the top of the table.
        \item \textit{Tools}: In each scene, we spawned $N \sim Uniform(5,20)$ tools which are selected randomly from $Uniform(1,49)$. The spawn tools are dropped onto the table from $[x,y,z]$ selected randomly from $Uniform(0, 1)$ above the table. Initial orientation (each dimension) is sampled from $Uniform(0,360)$.
    \end{itemize}
    The special cases when the sampled camera not seeing the table-top etc. are handled using hand-designed rules. Some examples from this process can be seen in Figure \ref{fig:synthetic_samples}.
    %For some tools such as \texttt{Grinder, Rake} the images collected from the web yielded insufficient examples, we formed a synthetic set as well. For this, we first photographed 10 background images that belonged to different contexts and manually labeled table-tops. Then, for each tool class for which we need more samples, we downloaded royalty-free transparent images, performed the following set of transformations:
    %\begin{itemize}
    %    \item \textit{Rotation}: Rotation around the center of the patch with an angle sampled from $Uniform(0, 2\pi)$.
    %    \item \textit{Scaling}: Scaling such that the longest dimension of the patch is between 12.5\% and 50\% percent of the area of the tabletop. 
    %    \item \textit{Shear transform} with a shear factor selected randomly from $Uniform(0.3, 1)$.
    %\end{itemize}
    %The number of tools added to an image is sampled from $Uniform(10,20)$.
\end{itemize}

For annotating the tools in the downloaded and the photographed images, we used the VGG Image Annotation (VIA) tool \cite{dutta2016via}. Annotation was performed by the authors of the paper.

\subsection{Dataset Statistics}

In this section, we provide some descriptive statistics about the METU-ALET dataset.

\begin{figure*}
    \centering
    \includegraphics[width=0.95\textwidth]{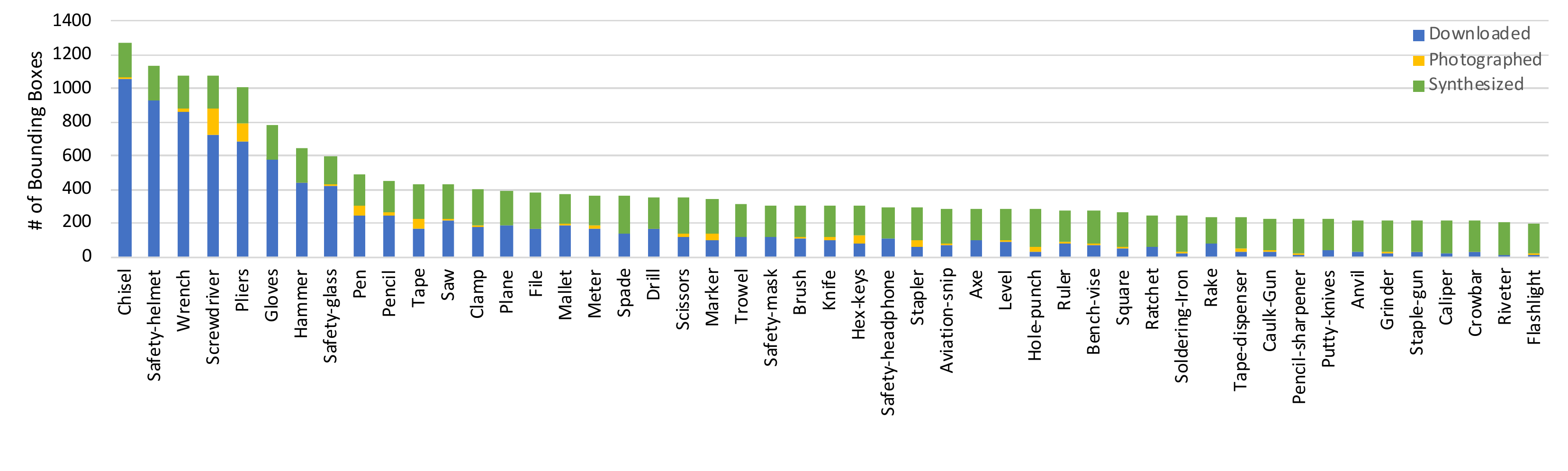}
    \vspace*{-0.8cm}
    \caption{The distribution of bounding boxes across classes. With the photographed and synthesized photos, each tool category has more than 200 bounding boxes. [Best viewed in color]\label{fig:class_BB_numbers}}
\end{figure*}

\textbf{Cardinality and Sizes of BBs}: The METU-ALET dataset includes 22,835 bounding boxes (BBs). As displayed in Fig. \ref{fig:class_BB_numbers}, \textit{for each tool category, there are more than 200 BBs, which is on an order similar to the widely used object detection datasets such as PASCAL \cite{PASCAL}}. As shown in Table \ref{tab:BB_sizes}, METU-ALET includes tools that appear small (area $<32^2$), medium ($32^2<$ area $<96^2$) and large ($96^2<$ area) -- following the naming convention from MS-COCO \cite{MSCOCO}. Although this is expected, as we will see in Section \ref{sect:results}, deep networks have difficulty especially detecting small tools. 

\begin{table}[]
    \centering
    \caption{The sizes of the bounding boxes (BB) of the annotated tools in METU-ALET. For calculating these statistics, we considered the scaled versions of the images that were fed to the networks, namely, $1333x800$.\label{tab:BB_sizes}}
    \scriptsize
    \begin{tabular}{|c|c|c|c|c|}\hline
        \CStart{} \textbf{Subset} \\ \textbf{Category}\CEnd{} &\textbf{ Small BBs} & \textbf{Medium BBs} & \textbf{Large BBs} & \textbf{Total} \\ \hline\hline
        Downloaded  & 809 & 4650 & 5661 & 11114\\\hline
        Photographed  & 13 & 309 & 443 & 765\\\hline
        Synthesized  & 813 & 6934 & 3209 & 10956 \\\hline
        Total & 1629 & 11893 & 9313 & 22835 \\\hline
    \end{tabular}
\end{table}

\textbf{Cardinality and Sizes of the Images}: Our dataset is composed of 2699 images in total, and on average, has size $1138\times 903$. See Table \ref{tab:image_statistics} for more details. Although the number of images may appear low, the number of bounding boxes (22835) is sufficient since there are more than 200 BBs per tool, and the avg. number of BBs per image is rather large ($6.6$) compared to PASCAL 2012 ($2.3$).

\begin{table}[]
    \centering
    \caption{The cardinality and the resolution of the images in METU-ALET. \label{tab:image_statistics}}
    \scriptsize
    \begin{tabular}{|c|c|c|}\hline
        \textbf{Subset} & \textbf{Cardinality} & \textbf{Avg. Resolution}  \\\hline\hline
        Downloaded & 1870 &  $924\times 786$\\\hline
        Photographed & 89 & $3663\times 3310$\\\hline
        Synthesized & 740 &  $1374\times 917$\\\hline
        Total/Avg & 2699 & $1138\times 903$\\\hline
    \end{tabular}
\end{table}

\subsection{ALET Safety Subset}

In this section, we illustrate how the ALET dataset can be used for addressing a critical issue in human-robot collaborative environments; that of checking whether the human worker is conforming to the security guidelines and wearing the required safety tools. ALET dataset contains a good number of safety tool (``helmet, glass, mask, headphone and glove'') instances ($2159$ in total just in real images, and $3104$ when combined with synthetic examples), which suffice for training a deep classifier. 

For forming the safety subset, first we used OpenPose \cite{OpenPose} to both detect the humans in an image and estimate their 2D poses. Then, we compared the positions of safety tools with the positions of the corresponding joints (e.g. for glove, the hand joints are used whereas head joints are used for the helmet etc.). If, for a tool, the Euclidean distance between the positions (normalized wrt. size of the BB of the human) were less than 0.2, the human was considered wearing that safety tool.

\section{METHODOLOGY}

In this section, we briefly describe the deep object detectors that we evaluated as a baseline, and the safety as a straightforward usecase of our dataset.

\subsection{Deep Object Detectors}

As stated in Section \ref{sect:related_work}, top-down deep object detectors can be broadly analyzed in two categories: (i) single-stage detectors, and (ii) multi-stage detectors. To form a baseline, we evaluated strong representatives of both single-stage (RetinaNet) and multi-stage (Faster R-CNN, Cascade R-CNN) detectors. Moreover, we included RepPoints, a recent bottom-up object detector.

\textbf{Faster R-CNN.} Faster R-CNN \cite{faster_RCNN} is one of the first networks to use end-to-end learning for object detection. It feeds features extracted from a backbone network to a region proposal network, which estimates an objectness score and the (relative) coordinates of a set of $k$ anchor boxes for each position on a regular grid. For each such box with an objectness score above a threshold, the object classification network (Fast R-CNN) is executed to classify each box into one of the object categories. 

For training the network, classification loss and box-regression loss (to penalize the spatial mismatch between the detected box and the ground truth box) are combined. The box-regression loss is weighted with a constant ($\lambda$), which we selected as 1.0 as suggested by the paper.

\textbf{Cascade R-CNN.} Cascade R-CNN \cite{Cascade} utilizes a region-proposal network with multiple detection stages with an increasing amount of IoU threshold for each of them. By doing so it eliminates negative samples better in each detection stage while increasing its IoU threshold in each further detection stage. 

%combines region-proposal and classification stages by making classification for a fixed set of default bounding boxes per position per object category. It is aimed to be a real-time object detector, and to achieve this, it assumes a coarser set of default boxes than similar networks such as SSD \cite{SSD}, leading to 9 default boxes per object category.

%For training the network, YOLO uses squared-error loss for both classification and box-regression to make learning easier to optimize and introduces several hyperparameters ($\lambda_{coord}$ and $\lambda_{noobj}$) to weight the contribution of squared-error loss corresponding to position, width, height, and classification. In our experiments, we take these factors as suggested by the paper.

\textbf{RetinaNet.} RetinaNet \cite{RetinaNet} is a one-stage detector which forms a multi-scale pyramid from the features obtained from the backbone network and performs classification and bounding box estimation in parallel for each layer (scale) of the pyramid. In order to address the data imbalance problem that affects single-stage detectors owing to background, RetinaNet proposes using \textit{focal loss} that decreases the contribution of the ``easy'' examples to the overall loss. Compared to other single-stage detectors, RetinaNet considers a denser set of bounding boxes to classify.

\textbf{RepPoint Detection.} Contrary to other approaches, RepPoint \cite{DBLP:RepPoint} is an anchor free, bottom-up detector. It is based on identifying representative points on objects and then combining these points into bounding boxes.

\subsection{A Safety Usecase for ALET}
\label{sect:method:usecase}

We created a CNN architecture consisting of three 2D convolutional layers and two fully connected layers. After each convolutional layer we added a batch normalization layer, and  each layer is also followed by ReLu activation. The final layer has five outputs with sigmoid activation. The network performs five-class (one for each safety tool) multi-label classification with binary cross-entropy. The network is trained on ALET Safety Dataset.

An alternative approach could be to combine the results of the tool detector and the pose detector. However, considering that the tool detection networks are having acceptable performance, we adopted an independent network for safety detection. Moreover, a tool detector would be detecting 49 tools in a scene 43 of which are irrelevant for our safety usecase. 

%\begin{table}[]
%    \centering
%    \caption{The CNN architecture used for the ``wearing helmet?'' usecase. After Conv layers, batch normalization is added. $W\times H\times C$ denotes the width, the height and the number of channels of a layer. \label{tab:CNN_arch}}
%    \begin{tabular}{|c|c|}\hline
%        \textbf{Layer}  &  \textbf{Description} \\ \hline\hline
%        Input & $100\times100\times3$ \\\hline
%        Conv1 & $100\times100\times16$\\\hline % F = 6, P=1, S = 2
%        Non-linearity  & ReLU \\\hline
%        Conv2 & $100\times100\times32$ \\\hline %F=3, P=1, S=2
%        Non-linearity  & ReLU \\\hline
%        
%        Conv3 & $100\times100\times64$ \\\hline %F=2, P=1, S=2
%        Non-linearity  & ReLU \\\hline
%        
%        FC & 640000 \\\hline
%        Non-linearity  & ReLU\\ \hline    
%        FC & 128\\ \hline
%        Non-linearity  & ReLU\\ \hline        
%        Output & 5\\ \hline
%    \end{tabular}
%\end{table}

\section{EXPERIMENTS}
\label{sect:results}

%In this section, we provide a baseline for the ALET dataset and show a practical usecase for such a dataset.

\subsection{Training and Implementation Details}
\label{sect:training_imp}

We split the ALET dataset (unless otherwise stated, this means real and synthetic images combined together) into $2112$ ($\%80$)  training, $268$ ($\%10$) validation and $264$ ($\%10$) testing samples. For training each network, the following libraries and settings are used (in all networks, the output layer is replaced with the tool categories, and the whole network except for the feature extracting backbone is updated during training): For each detector, the pre-trained network from mmdetection \cite{mmdetection} is used with backbone ResNet-50-FPN.

%\begin{itemize}
%    \item {Faster R-CNN}: The pre-trained network from mmdetection \cite{mmdetection} is used with backbone ResNet-50-FPN.
%    \item {Cascade R-CNN}: The pre-trained network from mmdetection \cite{mmdetection} is used with backbone ResNet-50-FPN.
%    \item \textbf{SSD}: 
%    \item {RetinaNet}: The pre-trained network from mmdetection \cite{mmdetection} is used with backbone ResNet-50-FPN.
%    \item {RepPoints}: The pre-trained network from mmdetection \cite{mmdetection} is used with backbone ResNet-50-FPN and moment-based bounding box conversion method utilized.
%\end{itemize}

\subsection{Quantitative Results for Tool Detection}

On the testing subset of ALET, we compare the performance of the detectors trained on the training subset of ALET. We use  the MS COCO style average precision (AP). %AP is a measure of area under the precision-recall curve, generally calculated in the literature  by averaging over the precision values at a discrete set of recall values \cite{PASCAL}:
%
%\begin{equation}
%    AP = \frac{1}{|\mathcal{R}|} \sum_{r \in \mathcal{R}} p(r),
%\end{equation}
%
%{\noindent}where $\mathcal{R}$ is the set of recall values considered, and $p(r)$ is the precision value at recall $r$. In our evaluations, we followed the PASCAL \cite{PASCAL} style and consider 11 different recalls for calculating average precision. mAP is then the mean of AP across categories.

Table \ref{tab:mAP} lists the AP and mAP values of the baseline networks on our dataset. We notice that the baseline detectors perform well on tools that are very distinctive and different from others; e.g. tape-dispenser, safety-helmet, hole-punch, pencil-sharpener. On such tools, we observe AP of up to 84.057 (Cascade R-CNN for pencil-sharpener). 

However, we see that the detectors have trouble in detecting especially tools that are too narrow, like pen-pencil, knife, file. There are several reasons for these results: (i) Object detectors have been designed for general object detection and need to extended to consider a wider range of anchors with a wider range of aspect ratios, which would increase the complexities of the networks. (ii) A second reason is that annotated boxes of small tools such as pen-pencil include more pixels of other objects than the annotated small object itself.
(iii) Moreover, tools such as screwdriver, chisel, file look very similar to each other. Moreover, these tools appear very similar to parts of other tools from a side view or from far (e.g., the front part of a drill is likely to be classified as a screwdriver, and in many cases, one half of a plier is detected as a chisel). 

These suggest that tool detection is indeed a very challenging problem especially owing to small tools and tools having very similar appearances to other tools. 

Table \ref{tab:Acc_diff} evaluates the performance improvement provided by the synthetic images from the simulation environment. We compared our dataset in with and without synthetic images by using RetinaNet and it shows that synthetic images elevate our dataset. The results for other networks are being produced and will be included in the final version.

\begin{table*}[]
    \centering
    \caption{MS COCO style AP of the baseline networks. Training and testing are both performed with the synthetic + real images.  \label{tab:mAP}}
    \scriptsize
    {\setlength{\tabcolsep}{0.23em}
    \begin{tabular}{|c|l|c|c|c|c|c|c|c|c|c|c|c|c|c|c|c|c|c|c|c|c|c|c|c|c|c|c|c|c|c|c|c|c|c|c|c|c|c|c|c|c|c|c|c|c|c|c|c|c|c|}\hline
             & 
            \rotatebox{90}{\textbf{Anvil}} &
            \rotatebox{90}{\textbf{Aviation-snip}} &
            \rotatebox{90}{\textbf{Axe}} &
            \rotatebox{90}{\textbf{Bench-vise}} &
            \rotatebox{90}{\textbf{Brush}} &
            \rotatebox{90}{\textbf{Caliper}} &
            \rotatebox{90}{\textbf{Caulk-Gun}} &
            \rotatebox{90}{\textbf{Chisel}} &
            \rotatebox{90}{\textbf{Clamp}} &
            \rotatebox{90}{\textbf{Crowbar}} &
            \rotatebox{90}{\textbf{Drill}} &
            \rotatebox{90}{\textbf{File}} &
            \rotatebox{90}{\textbf{Flashlight}} &
            \rotatebox{90}{\textbf{Gloves}} &
            \rotatebox{90}{\textbf{Grinder}} &
            \rotatebox{90}{\textbf{Hammer}} &
            \rotatebox{90}{\textbf{Hex-keys}} &
            \rotatebox{90}{\textbf{Hole-punch}} &
            \rotatebox{90}{\textbf{Knife}} &
            \rotatebox{90}{\textbf{Level}} &
            \rotatebox{90}{\textbf{Mallet}} &
            \rotatebox{90}{\textbf{Marker}} &
            \rotatebox{90}{\textbf{Meter}} &
            \rotatebox{90}{\textbf{Pen}} &
            \rotatebox{90}{\textbf{Pencil}} &
            \rotatebox{90}{\textbf{Pencil-sharpener}} &
            \rotatebox{90}{\textbf{Plane}} &
            \rotatebox{90}{\textbf{Pliers}} &
            \rotatebox{90}{\textbf{Putty-knives}} &
            \rotatebox{90}{\textbf{Rake}} &
            \rotatebox{90}{\textbf{Ratchet}} &
            \rotatebox{90}{\textbf{Riveter}} &
            \rotatebox{90}{\textbf{Ruler}} &
            \rotatebox{90}{\textbf{Safety-glass}} &
            \rotatebox{90}{\textbf{Safety-headphone}} &
            \rotatebox{90}{\textbf{Safety-helmet}} &
            \rotatebox{90}{\textbf{Safety-mask}} &
            \rotatebox{90}{\textbf{Saw}} &
            \rotatebox{90}{\textbf{Scissors}} &
            \rotatebox{90}{\textbf{Screwdriver}} &
            \rotatebox{90}{\textbf{Soldering-Iron}} &
            \rotatebox{90}{\textbf{Spade}} &
            \rotatebox{90}{\textbf{Square}} &
            \rotatebox{90}{\textbf{Staple-gun}} &
            \rotatebox{90}{\textbf{Stapler}} &
            \rotatebox{90}{\textbf{Tape}} &
            \rotatebox{90}{\textbf{Tape-dispenser}} &
            \rotatebox{90}{\textbf{Trowel}} &
            \rotatebox{90}{\textbf{Wrench}} &
            \rotatebox{90}{\textbf{mAP}}\\\hline
        F. R-CNN &  \rotatebox{90}{\ \ 72.459\ \ } & \rotatebox{90}{\ \ 10.055\ \ } & \rotatebox{90}{\ \ 32.406\ \ } & \rotatebox{90}{\ \ 52.399\ \ } & \rotatebox{90}{\ \ 10.387\ \ } & \rotatebox{90}{\ \ 39.164\ \ } & \rotatebox{90}{\ \ \textbf{29.794}\ \ } & \rotatebox{90}{\ \ 9.846\ \ } & \rotatebox{90}{\ \ 27.552\ \ } & \rotatebox{90}{\ \ 33.349\ \ } & \rotatebox{90}{\ \ 33.041\ \ } & \rotatebox{90}{\ \ 8.015\ \ } & \rotatebox{90}{\ \ 44.904\ \ } & \rotatebox{90}{\ \ 27.469\ \ } & \rotatebox{90}{\ \ \textbf{47.988}\ \ } & \rotatebox{90}{\ \ 19.544\ \ } & \rotatebox{90}{\ \ \textbf{21.196}\ \ } & \rotatebox{90}{\ \ 53.863\ \ } & \rotatebox{90}{\ \ 16.563\ \ } & \rotatebox{90}{\ \ 19.478\ \ } & \rotatebox{90}{\ \ 24.066\ \ } & \rotatebox{90}{\ \ 20.332\ \ } & \rotatebox{90}{\ \ 21.664\ \ } & \rotatebox{90}{\ \ 10.174\ \ } & \rotatebox{90}{\ \ 3.615\ \ } & \rotatebox{90}{\ \ 79.609\ \ } & \rotatebox{90}{\ \ 40.590\ \ } & \rotatebox{90}{\ \ 26.324\ \ } & \rotatebox{90}{\ \ 35.579\ \ } & \rotatebox{90}{\ \ 34.509\ \ } & \rotatebox{90}{\ \ \textbf{21.822}\ \ } & \rotatebox{90}{\ \ 19.221\ \ } & \rotatebox{90}{\ \ \textbf{6.183}\ \ } & \rotatebox{90}{\ \ 19.673\ \ } & \rotatebox{90}{\ \ 55.061\ \ } & \rotatebox{90}{\ \ 54.212\ \ } & \rotatebox{90}{\ \ 43.853\ \ } & \rotatebox{90}{\ \ 18.990\ \ } & \rotatebox{90}{\ \ 15.883\ \ } & \rotatebox{90}{\ \ 16.548\ \ } & \rotatebox{90}{\ \ 32.060\ \ } & \rotatebox{90}{\ \ 26.620\ \ } & \rotatebox{90}{\ \ 16.801\ \ } & \rotatebox{90}{\ \ 32.707\ \ } & \rotatebox{90}{\ \ 47.723\ \ } & \rotatebox{90}{\ \ \textbf{31.348}\ \ } & \rotatebox{90}{\ \ 44.904\ \ } & \rotatebox{90}{\ \ 12.473\ \ } & \rotatebox{90}{\ \ 23.657\ \ } & \rotatebox{90}{\ \ 29.504\ \ } \\ \hline
        Cascade &  \rotatebox{90}{\ \ \textbf{76.828}\ \ } & \rotatebox{90}{\ \ \textbf{12.192}\ \ } & \rotatebox{90}{\ \ \textbf{43.472}\ \ } & \rotatebox{90}{\ \ 55.545\ \ } & \rotatebox{90}{\ \ \textbf{13.906}\ \ } & \rotatebox{90}{\ \ \textbf{44.049}\ \ } & \rotatebox{90}{\ \ 29.496\ \ } & \rotatebox{90}{\ \ \textbf{11.990}\ \ } & \rotatebox{90}{\ \ \textbf{34.169}\ \ } & \rotatebox{90}{\ \ \textbf{42.046}\ \ } & \rotatebox{90}{\ \ 35.007\ \ } & \rotatebox{90}{\ \ \textbf{11.296}\ \ } & \rotatebox{90}{\ \ \textbf{51.400}\ \ } & \rotatebox{90}{\ \ \textbf{28.866}\ \ } & \rotatebox{90}{\ \ 47.635\ \ } & \rotatebox{90}{\ \ \textbf{23.255}\ \ } & \rotatebox{90}{\ \ 18.630\ \ } & \rotatebox{90}{\ \ \textbf{60.845}\ \ } & \rotatebox{90}{\ \ \textbf{19.880}\ \ } & \rotatebox{90}{\ \ \textbf{26.841}\ \ } & \rotatebox{90}{\ \ \textbf{33.148}\ \ } & \rotatebox{90}{\ \ \textbf{21.388}\ \ } & \rotatebox{90}{\ \ \textbf{27.889}\ \ } & \rotatebox{90}{\ \ \textbf{10.981}\ \ } & \rotatebox{90}{\ \ \textbf{8.870}\ \ } & \rotatebox{90}{\ \ \textbf{84.057}\ \ } & \rotatebox{90}{\ \ \textbf{45.020}\ \ } & \rotatebox{90}{\ \ \textbf{26.691}\ \ } & \rotatebox{90}{\ \ \textbf{35.582}\ \ } & \rotatebox{90}{\ \ \textbf{43.111}\ \ } & \rotatebox{90}{\ \ 19.815\ \ } & \rotatebox{90}{\ \ \textbf{26.048}\ \ } & \rotatebox{90}{\ \ 3.957\ \ } & \rotatebox{90}{\ \ \textbf{25.455}\ \ } & \rotatebox{90}{\ \ \textbf{58.154}\ \ } & \rotatebox{90}{\ \ \textbf{56.597}\ \ } & \rotatebox{90}{\ \ 46.912\ \ } & \rotatebox{90}{\ \ \textbf{20.759}\ \ } & \rotatebox{90}{\ \ \textbf{19.303}\ \ } & \rotatebox{90}{\ \ \textbf{21.622}\ \ } & \rotatebox{90}{\ \ \textbf{38.022}\ \ }  & \rotatebox{90}{\ \ \textbf{29.994}\ \ } & \rotatebox{90}{\ \ 18.053\ \ } & \rotatebox{90}{\ \ \textbf{43.114}\ \ } & \rotatebox{90}{\ \ \textbf{51.709}\ \ } & \rotatebox{90}{\ \ 30.281\ \ } & \rotatebox{90}{\ \ \textbf{52.868}\ \ } & \rotatebox{90}{\ \ \textbf{19.746}\ \ } & \rotatebox{90}{\ \ \textbf{29.843}\ \ } & \rotatebox{90}{\ \ \textbf{33.395}\ \ } \\ \hline
        RepPoint & \rotatebox{90}{\ \ 74.244\ \ } & \rotatebox{90}{\ \ 8.378\ \ } & \rotatebox{90}{\ \ 37.624\ \ } & \rotatebox{90}{\ \ \textbf{58.743}\ \ } & \rotatebox{90}{\ \ 13.565\ \ } & \rotatebox{90}{\ \ 40.194\ \ } & \rotatebox{90}{\ \ 19.457\ \ } & \rotatebox{90}{\ \ 10.765\ \ } & \rotatebox{90}{\ \ 27.696\ \ } & \rotatebox{90}{\ \ 32.818\ \ } & \rotatebox{90}{\ \ \textbf{35.832}\ \ } & \rotatebox{90}{\ \ 5.226\ \ } & \rotatebox{90}{\ \ 40.098\ \ } & \rotatebox{90}{\ \ 26.825\ \ } & \rotatebox{90}{\ \ 43.463\ \ } & \rotatebox{90}{\ \ 17.260\ \ } & \rotatebox{90}{\ \ 18.407\ \ } & \rotatebox{90}{\ \ 57.106\ \ } & \rotatebox{90}{\ \ 18.337\ \ } & \rotatebox{90}{\ \ 22.725\ \ } & \rotatebox{90}{\ \ 23.849\ \ } & \rotatebox{90}{\ \ 18.882\ \ } & \rotatebox{90}{\ \ 20.667\ \ } & \rotatebox{90}{\ \ 5.018\ \ } & \rotatebox{90}{\ \ 7.078\ \ } & \rotatebox{90}{\ \ 77.734\ \ } & \rotatebox{90}{\ \ 35.258\ \ } & \rotatebox{90}{\ \ 26.483\ \ } & \rotatebox{90}{\ \ 24.412\ \ } & \rotatebox{90}{\ \ 34.455\ \ } & \rotatebox{90}{\ \ 15.143\ \ } & \rotatebox{90}{\ \ 14.223\ \ } & \rotatebox{90}{\ \ 3.389\ \ } & \rotatebox{90}{\ \ 21.542\ \ } & \rotatebox{90}{\ \ 52.697\ \ } & \rotatebox{90}{\ \ 53.013\ \ } & \rotatebox{90}{\ \ 48.066\ \ } & \rotatebox{90}{\ \ 20.072\ \ } & \rotatebox{90}{\ \ 11.207\ \ } & \rotatebox{90}{\ \ 15.918\ \ } & \rotatebox{90}{\ \ 28.515\ \ } & \rotatebox{90}{\ \ 22.724\ \ } & \rotatebox{90}{\ \ \textbf{18.675}\ \ } & \rotatebox{90}{\ \ 25.806\ \ } & \rotatebox{90}{\ \ 42.978\ \ } & \rotatebox{90}{\ \ 26.009\ \ } & \rotatebox{90}{\ \ 45.661\ \ } & \rotatebox{90}{\ \ 10.115\ \ } & \rotatebox{90}{\ \ 22.828\ \ } & \rotatebox{90}{\ \ 28.187\ \ } \\ \hline
        RetinaNet &  \rotatebox{90}{\ \ 74.221\ \ } & \rotatebox{90}{\ \ 3.688\ \ } & \rotatebox{90}{\ \ 25.142\ \ } & \rotatebox{90}{\ \ 52.970\ \ } & \rotatebox{90}{\ \ 2.825\ \ } & \rotatebox{90}{\ \ 31.939\ \ } & \rotatebox{90}{\ \ 10.034\ \ } & \rotatebox{90}{\ \ 7.690\ \ } & \rotatebox{90}{\ \ 24.473\ \ } & \rotatebox{90}{\ \ 23.042\ \ } & \rotatebox{90}{\ \ 30.572\ \ } & \rotatebox{90}{\ \ 4.764\ \ } & \rotatebox{90}{\ \ 32.114\ \ } & \rotatebox{90}{\ \ 25.463\ \ } & \rotatebox{90}{\ \ 44.273\ \ } & \rotatebox{90}{\ \ 14.123\ \ } & \rotatebox{90}{\ \ 17.078\ \ } & \rotatebox{90}{\ \ 55.004\ \ } & \rotatebox{90}{\ \ 10.947\ \ } & \rotatebox{90}{\ \ 19.545\ \ } & \rotatebox{90}{\ \ 19.147\ \ } & \rotatebox{90}{\ \ 13.013\ \ } & \rotatebox{90}{\ \ 15.904\ \ } & \rotatebox{90}{\ \ 2.025\ \ } & \rotatebox{90}{\ \ 4.086\ \ } & \rotatebox{90}{\ \ 76.220\ \ } & \rotatebox{90}{\ \ 29.780\ \ } & \rotatebox{90}{\ \ 20.118\ \ } & \rotatebox{90}{\ \ 20.658\ \ } & \rotatebox{90}{\ \ 31.445\ \ } & \rotatebox{90}{\ \ 11.851\ \ } & \rotatebox{90}{\ \ 11.881\ \ } & \rotatebox{90}{\ \ 1.088\ \ } & \rotatebox{90}{\ \ 16.012\ \ } & \rotatebox{90}{\ \ 53.298\ \ } & \rotatebox{90}{\ \ 51.680\ \ } & \rotatebox{90}{\ \ \textbf{48.647}\ \ } & \rotatebox{90}{\ \ 8.440\ \ } & \rotatebox{90}{\ \ 11.380\ \ } & \rotatebox{90}{\ \ 8.178\ \ } & \rotatebox{90}{\ \ 22.776\ \ } & \rotatebox{90}{\ \ 11.488\ \ } & \rotatebox{90}{\ \ 13.854\ \ } & \rotatebox{90}{\ \ 30.117\ \ } & \rotatebox{90}{\ \ 46.282\ \ } & \rotatebox{90}{\ \ 27.571\ \ } & \rotatebox{90}{\ \ 43.356\ \ } & \rotatebox{90}{\ \ 6.855\ \ } & \rotatebox{90}{\ \ 11.294\ \ } & \rotatebox{90}{\ \ 24.048\ \ } \\ \hline
    \end{tabular}
    }

\end{table*}

\begin{table}[]
    \centering
    \caption{The gain obtained for RetinaNet using the synthetic examples generated from our simulation environment.  \label{tab:Acc_diff}}
    \scriptsize
    \begin{tabular}{|c|c|c|}\hline
          &  \textbf{Training w} & \textbf{Training w}   \\
          &   \textbf{Real Data} & \textbf{Real+Synthetic Data} \\
          \hline\hline
        mAp on Real+Syn test images & $11.980$ & $20.118$ \\ \hline
    \end{tabular}
\end{table}

\subsection{Sample Tool Detection Results}

\begin{figure}[hbt!]
    \centering
    \includegraphics[width=0.49\textwidth]{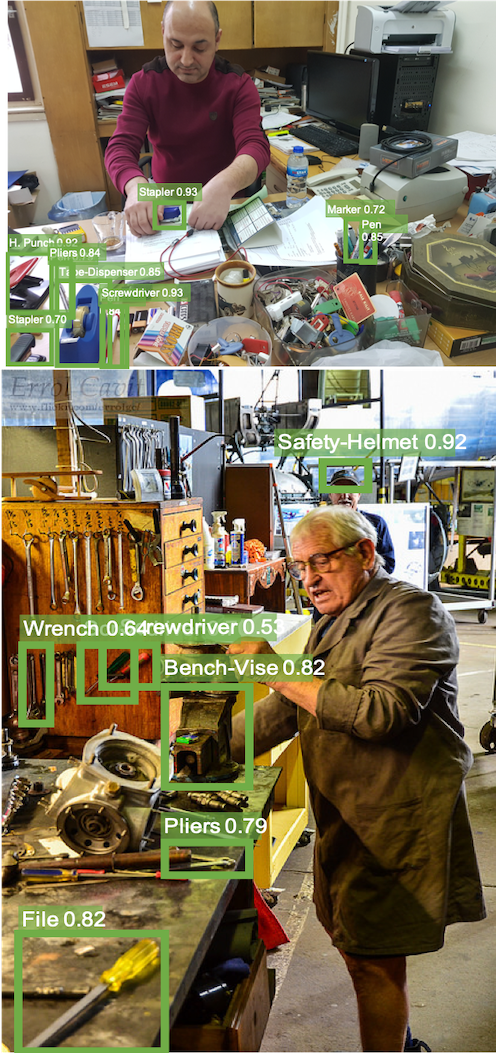}

    \caption{Sample detections from RetinaNet (w 0.4 as confidence threshold) on a few of the challenging scenes from Fig. \ref{fig:dataset_samples}. In the top example, almost all detections are correct. However, detector detected screwdriver and pliers in place of pens, which can be attributed to similarity in size and color. In the second image detector wrongfully annotates a hat as safety-helmet. In addition, it misses some of the crowded wrench examples, which is caused by the complexity of the problem.  \label{fig:example_detections}}
\end{figure}

In Fig. \ref{fig:example_detections}, we display a few detection results on a few of the challenging scenes in Fig.  \ref{fig:dataset_samples}. We see that although many tools are detected, many are missed.

\subsection{Safety Usecase Results}

\begin{figure}
    \centerline{
        \includegraphics[width=0.49\textwidth]{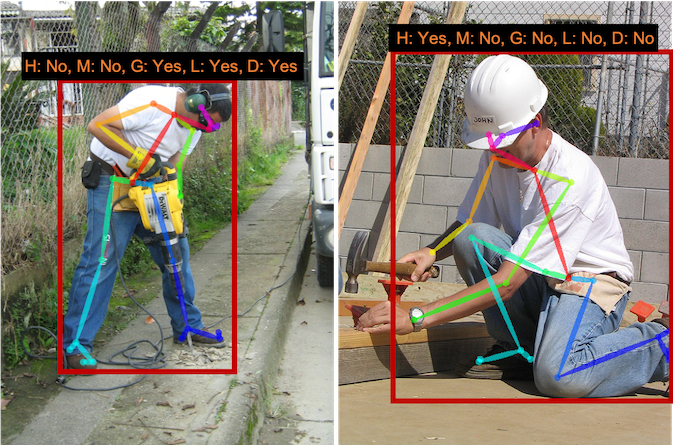}
    }
    \caption{Sample results for the safety usecase. Symbols: H: Helmet, M: Mask, G: Gloves, L: Glass, D: Headphone. [Best viewed in color]   \label{fig:wearing_helmet_results}}
\end{figure}

In this usecase, we analyze the method proposed in Section \ref{sect:method:usecase}. Our CNN network obtained $0.50$ F1 score on the test set, and as illustrated with some examples in Fig. \ref{fig:wearing_helmet_results} and accuracy results from Table \ref{tab:Class_Acc_Usecase}, we see that a network and a human detector \& pose estimator can be used to easily identify whether a human is wearing a safety equipment or not. However, the network is having difficulty with safety glasses and gloves, because, as illustrated in Fig. \ref{fig:wearing_helmet_results}, the region around eyes may not be visible for a working man and hands tend to be outside the BB in some cases.

\begin{table}[]
    \centering
    \caption{Class accuracies for the safety network. \label{tab:Class_Acc_Usecase}}
    \scriptsize
    \begin{tabular}{|c|c|c|c|c|c|}\hline
          &  \textbf{Helmet} & \textbf{Gloves} & \textbf{Mask} & \textbf{Headphone} & \textbf{Glass} \\ \hline\hline
        Accuracy & $75\%$ & $69\%$ & $85\%$ & $85\%$ & $54\%$\\ \hline
    \end{tabular}
\end{table}
\section{CONCLUSION}

In this paper, we have introduced an extensive dataset for tool detection in the wild. Moreover, we formed a baseline by training and testing four widely-used state-of-the-art deep object detectors in the literature, namely, Faster R-CNN \cite{faster_RCNN}, Cascade R-CNN \cite{Cascade}, and RetinaNet \cite{RetinaNet} and RepPoint Detection \cite{DBLP:RepPoint}. We demonstrated that such detectors especially have trouble in finding tools whose appearance is highly affected by viewpoint changes and tools that resemble parts of other tools.

Moreover, we have provided a very practical yet critical usecase for human-robot collaborative scenarios. Combining the detected ``helmet, glass, mask, headphone and gloves'' categories with the detection results of a human detector \& pose estimator, we have demonstrated how our dataset can be used for practical applications other than merely detecting tools in an environment.

\section*{ACKNOWLEDGMENT}
This work was supported by the Scientific and Technological Research Council of Turkey (T\"UB\.ITAK) through project called ``CIRAK: Compliant robot manipulator support for montage workers in factories'' (project no 117E002). The numerical calculations reported in this paper were partially performed at T\"UB\.ITAK ULAKBIM, High Performance and Grid Computing Center (TRUBA resources). We would like to thank Erfan Khalaji for his contributions on an earlier version of the work.

\bibliographystyle{IEEEtran}
\bibliography{references}

\end{document}